%% file: main.tex
\pdfoutput=1

\documentclass[11pt]{article}

\usepackage{naacl2021}

\usepackage{times}
\usepackage{latexsym}

\usepackage[T1]{fontenc}

\usepackage[utf8]{inputenc}

\usepackage{microtype}

\usepackage{amsfonts,amsmath,bm,xspace}
\usepackage{graphicx}
\usepackage{subcaption}
\usepackage{booktabs}
\usepackage{multicol,multirow}

\DeclareMathOperator*{\argmax}{arg\,max}
\DeclareMathOperator*{\argmin}{arg\,min}

\newcommand{\softmax}{\operatorname{softmax}}
\newcommand{\x}{\bm x}
\newcommand{\y}{\bm y}
\newcommand{\z}{\bm z}
\newcommand{\h}{\bm h}
\newcommand{\Q}{\mathbf{Q}}

\newcommand{\hit}{\checkmark}
\newcommand{\method}{\textsc{CNAT}\xspace}

\title{Non-Autoregressive Translation by Learning Target Categorical Codes}

\author{Yu Bao$^{\dag\star}$\quad Shujian Huang$^{\dag\star}$\quad Tong Xiao$^{\ddag}$\quad Dongqi Wang$^{\dag\star}$\\
\textbf{Xin-Yu Dai$^{\dag\star}$\quad Jiajun Chen$^{\dag\star}$}\\
$^{\dag}$National Key Laboratory for Novel Software Technology, Nanjing University \\ $^{\star}$Collaborative Innovation Center of Novel Software Technology and Industrialization\\
$^{\ddag}$NiuTrans Co., Ltd., Shenyang, China\\
\texttt{\{baoy,wangdq\}@smail.nju.edu.cn}, \texttt{xiaotong@mail.neu.edu.cn} \\
\texttt{\{huangsj,daixinyu,chenjj\}@nju.edu.cn}
}

\begin{document}
\maketitle

\input{00abstract}
\input{01introduction}
\input{02background}
\input{03method}

\input{04experiments}

\input{05related}
\input{06conclusion}


\bibliography{anthology,cnat}
\bibliographystyle{acl_natbib}

\appendix
\input{07appendix}
\end{document}

%% file: 00abstract.tex
\begin{abstract}
Non-autoregressive Transformer is a promising text generation model. 
However, current non-autoregressive models still fall behind their autoregressive counterparts in translation quality. 
We attribute this accuracy gap to the lack of dependency modeling among decoder inputs. 
In this paper, we propose \method, which learns implicitly categorical codes as latent variables into the non-autoregressive decoding. 
The interaction among these categorical codes remedies the missing dependencies and improves the model capacity.
Experiment results show that our model achieves comparable or better performance in machine translation tasks, compared with several strong baselines.
\end{abstract}

%% file: 01introduction.tex
\section{Introduction}
Non-autoregressive Transformer~\cite[NAT,][]{nat,nat_reg,iter_nat,cmlm} is a promising text generation model for machine translation. 
It introduces the conditional independent assumption among the target language outputs and simultaneously generates the whole sentence, bringing in a remarkable efficiency improvement~(more than $10\times$ speed-up) versus the autoregressive model. 
However, the NAT models still lay behind the autoregressive models in terms of BLEU~\citep{bleu} for machine translation. 
We attribute the low-quality of NAT models to the lack of dependencies modeling for the target outputs, making it harder to model the generation of the target side translation. 

A promising way is to model the dependencies of the target language by the latent variables. 
A line of research works~\cite{lt,vqvae,lv_nar,flowseq} introduce latent variable modeling to the non-autoregressive Transformer and improves translation quality. 
The latent variables could be regarded as the springboard to bridge the modeling gap, introducing more informative decoder inputs than the previously copied inputs. 
More specifically, the latent-variable based model first predicts a latent variable sequence conditioned on the source representation, where each variable represents a chunk of words. The model then simultaneously could generate all the target tokens conditioning on the latent sequence and the source representation since the target dependencies have been modeled into the latent sequence.

However, due to the modeling complexity of the chunks, the above approaches always rely on a large number~(more than $2^{15}$,~\citealp{lt,vqvae}) of latent codes for discrete latent spaces, which may hurt the translation efficiency---the essential goal of non-autoregressive decoding.


~\citet{syn_st} introduce syntactic labels as a proxy to the learned discrete latent space and improve the NATs' performance.  
The syntactic label greatly reduces the search space of latent codes, leading to a better performance in both quality and speed. However, it needs an external syntactic parser to produce the reference syntactic tree, which may only be effective in limited scenarios.
Thus, it is still challenging to model the dependency between latent variables for non-autoregressive decoding efficiently.

In this paper, we propose to learn a set of latent codes that can act like the syntactic label, which is learned without using the explicit syntactic trees. 
To learn these codes in an unsupervised way, we use each latent code to represent a fuzzy target category instead of a chunk as the previous research~\cite{syn_st}.
More specifically, we first employ vector quantization~\cite{vqvae} to discretize the target language to the latent space with a smaller number~(less than 128) of latent variables, which can serve as the fuzzy word-class information each target language word.
We then model the latent variables with conditional random fields~\cite[CRF,][]{crf,nat_crf}.
To avoid the mismatch of the training and inference for latent variable modeling, we propose using a gated neural network to form the decoder inputs. 
Equipping it with scheduled sampling~\cite{bengio2015scheduled}, the model works more robustly.

Experiment results on WMT14 and IWSLT14 show that \method achieves the new state-of-the-art performance without knowledge distillation. 
With the sequence-level knowledge distillation and reranking techniques, the \method is comparable to the current state-of-the-art iterative-based model while keeping a competitive decoding speedup.


%% file: 02background.tex
\section{Background}\label{sec:background}
Neural machine translation~(NMT) is formulated as a conditional probability model $p(\y|\x)$, which models a sentence $\y = \{y_1,y_2,\cdots,y_m\}$ in the target language given the input $\x = \{x_1,x_2,\cdots,x_n\}$ from the source language.



\subsection{Non-Autoregressive Neural Machine Translation}\label{ss:non-auto}
\citet{nat} proposes Non-Autoregressive Transformer~(NAT) for machine translation, breaking the dependency among target tokens, thus achieving simultaneous decoding for all tokens. 
For a source sentence, a non-autoregressive decoder factorizes the probability of its target sentence as:
\begin{equation}
    p(\y|\x)= \prod_{t=1}^{m}p(y_{t}|\x; \theta),
    \label{eqn:nat}
\end{equation}
where $\theta$ is the set of model parameters.

\begin{figure}[tbp]
    \centering
    \small
    \begin{minipage}{0.3\linewidth}
    \centering
    \includegraphics[width=\linewidth]{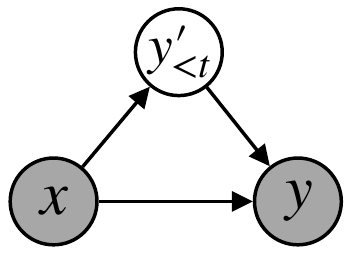}
    \subcaption{AT}
    \label{fig:at}
    \end{minipage}
    \hfill
    \begin{minipage}{0.3\linewidth}
    \centering
    \includegraphics[width=\linewidth]{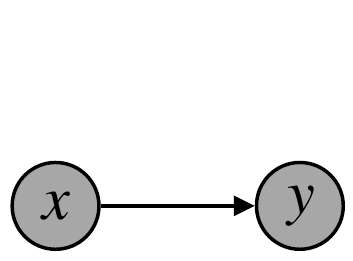}
    \subcaption{NAT}
    \label{fig:nat}
    \end{minipage}
    \hfill
    \begin{minipage}{0.3\linewidth}
    \centering
    \includegraphics[width=\linewidth]{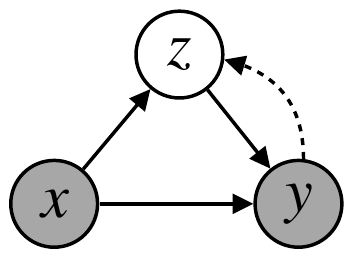}
    \subcaption{LT}
    \label{fig:lt}
    \end{minipage}
\caption{Different inference process of different Transformer models.}
\label{fig:lm}
\end{figure}

NAT has a similar architecture to the autoregressive Transformer~\citep[AT,][]{transformer}, which consists of a multi-head attention based encoder and decoder. 
The model first encodes the source sentence $x_{1:n}$ as the contextual representation $e_{1:n}$, then employs an extra module to predict the target length and form the decoder inputs.
\begin{itemize}
    \item \textbf{Length Prediction: } Specifically, the length predictor in the bridge module predicts the target sequence length $m$ by:
    \begin{equation}
        m = n + \argmax_{\Delta L} p(\Delta_{L}|\operatorname{mean}(e); \phi),
    \end{equation}
    where $\Delta_{L}$ is the length difference between the target and source sentence, $\phi$ is the parameter of length predictor.
    \item \textbf{Inputs Initialization: } With the target sequence length $m$, we can compute the decoder inputs $\h=h_{1:m}$ with \textit{Softcopy}~\cite{hint_nat,imitate_nat} as:
    \begin{equation}
        \begin{split}
          h_j & = \sum_{i}^{n} w_{ij}\cdot e_i  \\
        \text{and}\ w_{ij} & = \softmax (-|j-i|/\tau),
    \end{split}
    \label{equ:softcopy}
    \end{equation}
    where $\tau$ is a hyper-parameter to control the sharpness of the $\softmax$ function.
\end{itemize}
With the computed decoder inputs $h$, NAT generates target sequences simultaneously by $\argmax_{y_t}\ p(y_t|x; \theta)$ for each timestep $t$, effectively reduce computational overhead in decoding~(see Figure~\ref{fig:nat}). 

Though NAT achieves around $10\times$ speedup in machine translation than autoregressive models, it still suffers from potential performance degradation~\citep{nat}.
The results degrade since the removal of target dependencies prevents the decoder from leveraging the inherent sentence structure in prediction.
Moreover, taking the copied source representation as decoder inputs implicitly assume that the source and target language share a similar order, which may not always be the case~\cite{pnat}.

\subsection{Latent Transformer}\label{ss:latent-nat}
To bridge the gap between non-autoregressive and autoregressive decoding, \citet{lt} introduce the Latent Transformer~(LT).
It incorporates non-autoregressive decoding with conditional dependency as the latent variable to alleviate the degradation resulted from the absence of dependency:
\begin{equation}
\label{eqn:lt}
    p(\y|\x)= p(\z|\x;\phi) \prod_{t=1}^{m} p(y_{t}|\z,\x; \theta),
\end{equation}
where $\z=\{z_1,\cdots,z_L\}$ is the latent sequence and the $L$ is the length of the latent sequence, $\phi$ and $\theta$ are the parameter of latent predictor and translation model, respectively. 

The LT architecture stays unchanged from the origin NAT models, except for the latent predictor and decoder inputs. 
During inference, the Latent Transformer first autoregressively predicts the latent variables $\z$, then non-autoregressively produces the entire target sentence $\y$ conditioned on the latent sequence $\z$~(see Figure~\ref{fig:lt}). \citet{flowseq,lv_nar} extend this idea and model $\z$ as the continuous latent variables, achieving a promising result, which replaces the autoregressive predictor with the iterative transformation layer.


%% file: 03method.tex
\section{Approach}
In this section, we present our proposed~\method, an extension to the Transformer incorporated with non-autoregressive decoding for target tokens and autoregressive decoding for latent sequences. 

In brief, \method follows the architecture of Latent Transformer~\cite{lt}, except for the latent variable modeling~(in $\S$~\ref{ss:vq} and $\S$~\ref{ss:crf}) and inputs initialization~(in $\S$~\ref{ss:input}). 




\subsection{Modeling Target Categorical Information by Vector Quantization}\label{ss:vq}
Categorical information has achieved great success in neural machine translation, such as part-of-speech~(POS) tag in autoregressive translation~\cite{latent_pos} and syntactic label in non-autoregressive translation~\cite{syn_st}.

Inspired by the broad application of categorical information, we propose to model the implicit categorical information of target words in a non-autoregressive Transformer. 
Each target sequence $\y=y_{1:m}$ will be assigned to a discrete latent variable sequence $\z = z_{1:m}$. 
We assume that each $z_i$ will capture the fuzzy category of its token $y_i$.
Then, the conditional probability $p(\y|\x)$ is factorized with respect to the categorical latent variable:
\begin{equation}\label{eqn:pos}
    p(\y|\x) = \sum_{\z} p(\z | \x) \cdot p(\y|\z,\x).
\end{equation}
However, it is computationally intractable to sum all configurations of latent variables. 
Following the spirit of the latent based model~\cite{lt,vqvae}, we employ a vector quantized technique to maintain differentiability through the categorical modeling and learn the latent variables straightforward.

\paragraph{Vector Quantization.}
The vector quantization based methods have a long history of being successfully in machine learning models. 
In vector quantization, each target representation $\operatorname{repr}(y_i) \in \mathbb{R}^{d_\text{model}}$ is passed through a discretization bottleneck using a nearest-neighbor lookup on embedding matrix $\Q\in \mathbb{R}^{K \times d_\text{model}}$, where $K$ is the number of categorical codes.

For each $y_i$ in the target sequence, we define its categorical variable $z_i$ and latent code $q_i$ as:
\begin{equation}
\begin{split}
    z_i &= k,\ q_i = \Q_k, \\
\text{and}\ k & =\argmin_{j\in [K]} || \operatorname{repr}(y_i) - \Q_j ||_{2},
\end{split}
\label{eqn:vq}
\end{equation}
where $||\cdot||_{2}$ is the ${l}_2$ distance, $[K]$ denote the set $\{1,2,\cdots,K\}$.
Intuitively, we adopt the embedding of $\y$ as the target representation:
\begin{equation*}
    \operatorname{repr}(y_i) = \operatorname{embedding}(y_i)
\end{equation*}
where the embedding matrix of the target language is shared with the $\operatorname{softmax}$ layer of the decoder.

\paragraph{Exponential Moving Average.} 
Following the common practice of vector quantization, we also employ the exponential moving average~(EMA) technique to regularize the categorical codes. 

Put simply, the EMA technique could be understood as basically the k-means clustering of the hidden states with a sort of momentum.  
We maintain an EMA over the following two quantities for each $j \in [K]$: 1) the count $c_j$ measuring the number of target representations that have $\Q_j$ as its nearest neighbor, and 2) $\Q_j$. 
The counts are updated over a mini-batch of targets $\{y_1, y_2,\cdots,y_{m\times B}\}$ with:
\begin{equation}
    c_j  = \lambda c_j + (1-\lambda) \sum_{i}^{m\times B} \operatorname{1}[z_i=j],
    \label{eqn:ema_c}
\end{equation}
then, the latent code $\Q_j$ being updated with:
\begin{equation}
    \Q_j = \lambda \Q_j + (1-\lambda)\sum_{i}^{m\times B}\frac{\operatorname{1}[z_i=j]\operatorname{repr}(y_i)}{c_j},
    \label{eqn:ema_q}
\end{equation}
where $\operatorname{1}[\cdot]$ is the indicator function and $\lambda$ is a decay parameter, $B$ is the size of the batch.

\subsection{Modeling Categorical Sequence with Conditional Random Fields}\label{ss:crf}
Our next insight is transferring the dependencies among the target outputs into the latent spaces. 
Since the categorical variable captures the fuzzy target class information, it can be a proxy of the target outputs. 
We further employ a structural prediction module instead of the standard autoregressive Transformer to model the latent sequence. 
The former can explicitly model the dependencies among the latent variables and performs exact decoding during inference.

\paragraph{Conditional Random Fields.}
We employ a linear-chain conditional random fields~\cite[CRF,][]{crf} to model the categorical latent variables, which is the most common structural prediction model.

Given the source input $\x=(x_1,\cdots,x_n)$ and its corresponding latent variable sequence $\z=(z_1,\cdots,z_m)$, the CRF model defines the probability of $\z$ as:
\begin{equation}\begin{split}
    p(\z|\x)=\frac{1}{\mathbb{Z}(\x)}&\operatorname{exp}\Big(\sum_{i=1}^{m}s(z_i,\x,i) \\ 
    &+\sum_{i=2}^{m}t(z_{i-1},z_i,\x,i) \Big), 
    \label{eqn:crf}
\end{split}
\end{equation}
where $\mathbb{Z}(\x)$ is the normalize factor, $s(z_i,\x,i)$ is the emit score of $z_i$ at the position $i$, and the $t(z_{i-1},z_i,\x,i)$ is the transition score from $z_{i-1}$ to $z_i$. 

Before computing the emit score and transition score in Eq.~\ref{eqn:crf}, we first take $\h=h_{1:m}$ as the inputs and compute the representation $\bm f = \operatorname{Transfer}(\h)$, where $\operatorname{Transfer}(\cdot)$ denotes a two-layer vanilla Transformer decoding function including a self-attention block, an encoder-decoder block followed by a feed-forward neural network block~\cite{transformer}.

We then compute the emit score and the transition score. 
For each position $i$, we compute the emit score with a linear transformation: $s(z_i,\x,i)=(W^{T}f_{i}+b)_{z_i}$ where $W \in \mathbb{R}^{d_\text{model}\times K}$ and $b \in \mathbb{R}^{K}$ are the parameters. 
We incorporate the positional context and compute its transition score with:
\begin{equation}
    \begin{split}
        &\bm M_\text{d}^{i} = \operatorname{Biaffine}([f_{i-1};f_i]), \\
        &\bm M^{i}  = \bm E_{1}^{T}\bm M_\text{d}^{i}\bm E_2,\\
        &t(z_{i-1},z_i,\x,i) =\bm M_{z_{i-1},z_{i}}^{i},
    \end{split}
\end{equation}
where $\operatorname{Biaffine}(\cdot):\mathbb{R}^{2d_\text{model}} \to \mathbb{R}^{d_\text{t}\times d_\text{t}} $ is a biaffine neural network~\cite{biaffine}, $\bm E_1$ and $\bm E_2\in\mathbb{R}^{d_\text{t}\times K}$ are the transition matrix.

\subsection{Fusing Source Inputs and Latent Codes via Gated Function}\label{ss:input}
One potential issue is that the mismatch of the training and inference stage for the used categorical variables. 
Suppose we train the decoder with the quantized categorical variables $\z$, which is inferred from the target reference. 
In that case, we may fail to achieve satisfactory performance with the predicted categorical variables during inference.

We intuitively apply the gated neural network~(denote as \textbf{GateNet}) to form the decoder inputs by fusing the copied decoder inputs $\h=h_{1:m}$ and the latent codes $\bm q=q_{1:m}$, since the copied decoder inputs $\h$ is still informative to non-autoregressive decoding:
\begin{equation}
\begin{split}
    g_i &= \sigma(\operatorname{FFN}([h_i;q_i])), \\
     o_i &= h_i*g_i + q(z_i)*(1-g_i), 
\end{split}
\end{equation}
where the $\operatorname{FFN}(\cdot): \mathbb{R}^{2d_\text{model}}\to \mathbb{R}^{d_\text{model}}$ is a two-layer feed-forward neural networks and $\sigma(.)$ is the $\operatorname{sigmoid}$ function.
\subsection{Training}
While training, we first compute the reference $\z^\text{ref}$ by the vector quantization and employ the EMA to update the quantized codes.
The loss of the CRF-based predictor is computed with:
\begin{equation}
    \mathcal{L}_\text{crf} = -\operatorname{log} p(\z^\text{ref}|\x).
\end{equation}
To equip with the GateNet, we randomly mix the $\z^\text{ref}$ and the predicted $\z^\text{pred}$ as:
\begin{equation}
    \z_i^\text{mix}=\left\{
\begin{aligned}
\z_i^\text{pred} &\quad \text{if}\ p \geq \tau \\
\z_i^\text{ref\ \ \ } &\quad \text{if}\ p < \tau 
\end{aligned}
\right.,
\label{eqn:mix}
\end{equation}
where $p\sim \mathbb{U}[0,1]$ and $\tau$ is the threshold we set 0.5 in our experiments. 
Grounding on the $\z_\text{mix}$, the non-autoregressive translation loss is computed with: 
\begin{equation}
    \mathcal{L}_\text{NAT} =  -\operatorname{log} p(\y|\z^\text{mix},\x;\theta).
\end{equation}
With the hyper-parameter $\alpha$, the overall training loss is:
\begin{equation}
    \mathcal{L} = \mathcal{L}_\text{NAT} + \alpha\mathcal{L}_\text{crf}.
    \label{eqn:loss}
\end{equation}



\subsection{Inference} 
\method selects the best sequence by choosing the highest-probability latent sequence $z$ with \textit{Viterbi decoding}~\cite{viterbi1967error}, then generate the tokens with:
\begin{align*}
    \z^{*} &= \argmax_{\z} p(\z|\x;\theta), \\ 
    \text{and}\ \y^{*} &=\argmax_{\y}p(\y|\z^{*},\x;\theta),
\end{align*}
where identifying $\y^{*}$ only requires independently maximizing the local probability for each output position.

%% file: 04experiments.tex
\section{Experiments}\label{s:exp}
\paragraph{Datasets.} 
We conduct the experiments on the most widely used machine translation benchmarks: WMT14 English-German~(WMT14 EN-DE, 4.5M pairs)\footnote{\url{https://drive.google.com/uc?export=download&id=0B_bZck-ksdkpM25jRUN2X2UxMm8}} and IWSLT14 German-English~(IWSLT14, 160K pairs)\footnote{ \url{https://github.com/pytorch/fairseq}}. 
The datasets are processed with the Moses script~\citep{Moses}, and the words are segmented into subword units using byte-pair encoding~\citep[BPE]{bpe}.  
We use the shared subword embeddings between the source language and target language for the WMT datasets and the separated subword embeddings for the IWSLT14 dataset.
   
\paragraph{Model Setting.} 
In the case of IWSLT14 task, we use a small setting~($d_\text{model}$ = 256, $d_\text{hidden}$ = 512, $p_\text{dropout}$ = 0.1, $n_\text{layer}$ = 5 and $n_\text{head}$ = 4) for Transformer and NAT models. 
For the WMT tasks, we use the \texttt{Transformer-base} setting~($d_\text{model}$ = 512, $d_\text{hidden}$ = 512, $p_\text{dropout}$ = 0.3, $n_\text{head}$ = 8 and $n_\text{layer}$ = 6) of the~\citet{transformer}. 
We set the hyperparameter $\alpha$ used in Eq.~\ref{eqn:loss} and $\lambda$ in Eq.~\ref{eqn:ema_c}-\ref{eqn:ema_q} to 1.0 and 0.999, respectively.
The categorical number $K$ is set to 64 in our experiments.
We implement our model based on the open-source framework of \texttt{fairseq}~\cite{fairseq}.

\paragraph{Optimization.} 
We optimize the parameter with the Adam~\citep{adam} with $\beta=(0.9,0.98)$. 
We use inverse square root learning rate scheduling~\citep{transformer} for the WMT tasks and linear annealing schedule~\cite{iter_nat} from $3\times10^{-4}$ to $1\times10^{-5}$ for the IWSLT14 task.
Each mini-batch consists of 2048 tokens for IWSLT14 and 32K tokens for WMT tasks.

\paragraph{Distillation.}
Sequence-level knowledge distillation~\citep{hinton2015distilling} is applied to alleviate the multi-modality problem~\cite{nat} while training. 
We follow previous studies on NAT~\citep{nat,iter_nat,imitate_nat} and use translations produced by a pre-trained autoregressive Transformer~\cite{transformer} as the training data.

\paragraph{Reranking.}
We also include the results that come at reranked parallel decoding~\citep{nat,enat,nat_reg,imitate_nat}, which generates several decoding candidates in parallel and selects the best via re-scoring using a pre-trained autoregressive model. 
Specifically, we first predict the target length $\hat{m}$ and generate output sequence with $\arg\max$ decoding for each length candidate $m \in [\hat{m}-\Delta m, \hat{m}+\Delta m ]$~($\Delta m$ = 4 in our experiments, means there are $N=9$ candidates), which was called length parallel decoding~(LPD). 
Then, we use the pre-trained teacher to rank these sequences and identify the best overall output as the final output.

\paragraph{Baselines.}
We compare the \method with several strong NAT baselines, including:
\begin{itemize}
    \item The NAT builds upon latent variables: NAT-FT~\cite{nat}, LT~\cite{lt}, Syn-ST~\cite{syn_st}, LV-NAR~\cite{lv_nar} and Flowseq~\cite{flowseq}.
    \item The NAT with extra autoregressive decoding or iterative refinement: NAT-DCRF~\cite{nat_crf}, IR-NAT~\cite{iter_nat}, and CMLM~\cite{cmlm}.
    \item The NAT with auxiliary training objectives: NAT-REG~\cite{nat_reg}, imitate-NAT~\cite{imitate_nat}.
\end{itemize}
We compare the proposed \method against baselines both in terms of generating quality and inference speedup. 
For all our tasks, we obtain the performance of baselines by either directly using the performance figures reported in the previous works if they are available or producing them by using the open-source implementation of baseline algorithms on our datasets. 

\paragraph{Metrics.}
We evaluate using the tokenized and cased BLEU scores~\citep{bleu}. 
We highlight the best \textbf{NAT result} with bold text.

\subsection{Results}

\begin{table}[tbp]
\centering
\small
\begin{tabular}{lccc}
\toprule
\multirow{2}{*}{Model} & \multicolumn{2}{c}{WMT14} & IWSLT14\\
                       & EN-DE       & DE-EN     & DE-EN \\
\midrule
LV-NAR                 & 11.80       & /         & / \\
AXE CMLM               & 20.40       & 24.90     & / \\
SynST                  & 20.74       & 25.50     & 23.82\\  
Flowseq                & 20.85       & 25.40     & 24.75\\
\midrule
NAT~(ours)             & 9.80        & 11.02     & 17.77\\
\method~(ours)         & \textbf{21.30} & \textbf{25.73} & \textbf{29.81}\\
\bottomrule
\end{tabular}
\caption{Results of the NAT models with argmax decoding on test set of WMT14 and IWSLT14.}
\label{tab:pure_mt}
\end{table}

\paragraph{Translation Quality.}
First, we compare~\method with the NAT models without using advanced techniques, such as knowledge distillation, reranking, or iterative refinements.
The results are listed in Table~\ref{tab:pure_mt}.
The \method achieves significant improvements~(around 11.5 BLEU in EN-DE, more than 14.5 BLEU in DE-EN) over the vanilla NAT, which indicates that modeling categorical information could improve the modeling capability of the NAT model. 
Also, the \method achieves better results than Flowseq and SynST, which demonstrates the effectiveness of \method in modeling dependencies between the target outputs.

\begin{table}[tbp]
\centering
\small
\begin{tabular}{lccc}
\toprule
\multirow{2}{*}{Model}  & \multicolumn{2}{c}{WMT14} & IWSLT14 \\
                        & EN-DE         & DE-EN     & DE-EN \\
\midrule
NAT-FT                  & 17.69         & 21.47     &   /   \\
LT                      & 19.80         & /         &   /   \\
NAT-REG                 & 20.65         & 24.77     & 23.89 \\
imitate-NAT             & 22.44         & 25.67     &   /   \\
Flowseq                 & 23.72         & 28.39     & 27.55 \\
NAT-DCRF                &  23.44        & 27.22     & 27.44 \\
\midrule
Transformer~(ours)      & 27.33         & 31.69     & 34.29 \\
NAT~(ours)              & 17.69         & 18.93     & 23.78 \\
\method~(ours)          & \textbf{25.56}&\textbf{29.36} & \textbf{31.15}  \\
\bottomrule
\end{tabular}
\caption{Results of NAT models trained with knowledge distillation on test set of WMT14 and IWSLT14.}
\label{tab:kd_mt}
\end{table}

The performance of the NAT models with advance techniques~(sequence-level knowledge distillation or reranking) is listed in Table~\ref{tab:kd_mt} and Table~\ref{tab:lpd_mt}.
Coupling with the knowledge distillation techniques, all NAT models achieve remarkable improvements.

\begin{table}[tbp]
\centering
\small
\begin{tabular}{lccc}
\toprule
\multirow{2}{*}{Model}  & \multirow{2}{*}{N} & \multicolumn{2}{c}{WMT14} \\
                        &                    & EN-DE       & DE-EN      \\
\midrule
NAT-FT                  & 10                 & 18.66       & 22.42      \\
NAT-FT                  & 100                & 19.17       & 23.20      \\
LT                      & 10                 & 21.00       & /          \\
LT                      & 100                & 22.50       & /          \\
NAT-REG                 & 9                  & 24.61       & 28.90      \\
imitate-NAT             & 9                  & 24.15       & 27.28      \\
Flowseq                 & 15                 & 24.70       & 29.44      \\
Flowseq                 & 30                 & 25.31       & 30.68      \\
NAT-DCRF                & 9                  & 26.07       & 29.68      \\
NAT-DCRF                & 19                 & \textbf{26.80}       & 30.04      \\
\midrule
Transformer (ours)      & -                  & 27.33       & 31.69      \\
\method (ours)          & 9                  & 26.60       & \textbf{30.75}     \\
\bottomrule 
\end{tabular}
\caption{Results of NAT models with parallel decoding on test set of WMT14. ``N'' means the number of candidates to be re-ranked.}
\label{tab:lpd_mt}
\end{table}

Our best results are obtained with length parallel decoding, which employs a pretrained Transformer to rerank the multiple parallels generated candidates of different target lengths.
Specifically, on a large scale WMT14 dataset, \method surpasses the NAT-DCRF by 0.71 BLEU score in DE-EN but slightly under the NAT-DCRF around 0.20 BLEU in EN-DE, which shows that the~\method is comparable to the state-of-the-art NAT model.
Also, we can see that a larger ``N'' leads to better results~($N=100$ vs. $N=10$ of NAT-FT, $N=19$ vs. $N=9$ of NAT-DCRF, etc.); however, it always comes at the degradation of decoding efficiency.

\begin{table}[tbp]
\centering
\tabcolsep 4pt
\small
\begin{tabular}{lcccc}
\toprule
\multirow{2}{*}{Model}      & \multirow{2}{*}{Iteration} & \multicolumn{3}{c}{WMT14} \\
                            &                            & EN-DE         & DE-EN     & Speedup\\
\midrule
\multirow{4}{*}{IR-NAT}     & 1                          & 13.91         & 16.77     & 11.39$\times$ \\
                            & 2                          & 16.95         & 20.39     & 8.77$\times$\\
                            & 5                          & 20.26         & 23.86     & 3.11$\times$\\
                            & 10                         & 21.61         & 25.48     & 2.01$\times$\\
\midrule
\multirow{2}{*}{CMLM}  & 4                          & 26.08         & 30.11     &  / \\
                            & 10                         & \textbf{26.92}& \textbf{30.86}&  /   \\
\midrule
\method              & 1                          & 25.56         & 29.36        & 10.37$\times$ \\
\method (N=9)         & 1                          & 26.60         & 30.75        & 5.59$\times$ \\
\bottomrule
\end{tabular}
\caption{Results of NAT models with iterative refinements on test set of WMT14. ``Iteration'' means the number of iteration refinements.}
\label{tab:iter_mt}
\end{table}

We also compare our \method with the NAT models that employ an iterative decoding technique and list the results in Table~\ref{tab:iter_mt}.
The iterative-based non-autoregressive Transformer captures the target language's dependencies by iterative generating based on the previous iteration output, which is an important exploration for a non-autoregressive generation. 
With the iteration number increasing, the performance improving, the decoding speed-up dropping,  whatever the IR-NAT or CMLM.
We can see that the \method achieves a better result than the CMLM with four iterations and IR-NAT with ten iterations, even close to the CMLM with ten iterations while keeping the benefits of a one-shot generation. 

\begin{figure*}[tbp]
    \begin{subfigure}[t]{0.495\linewidth}
    \includegraphics[width=1.0\linewidth]{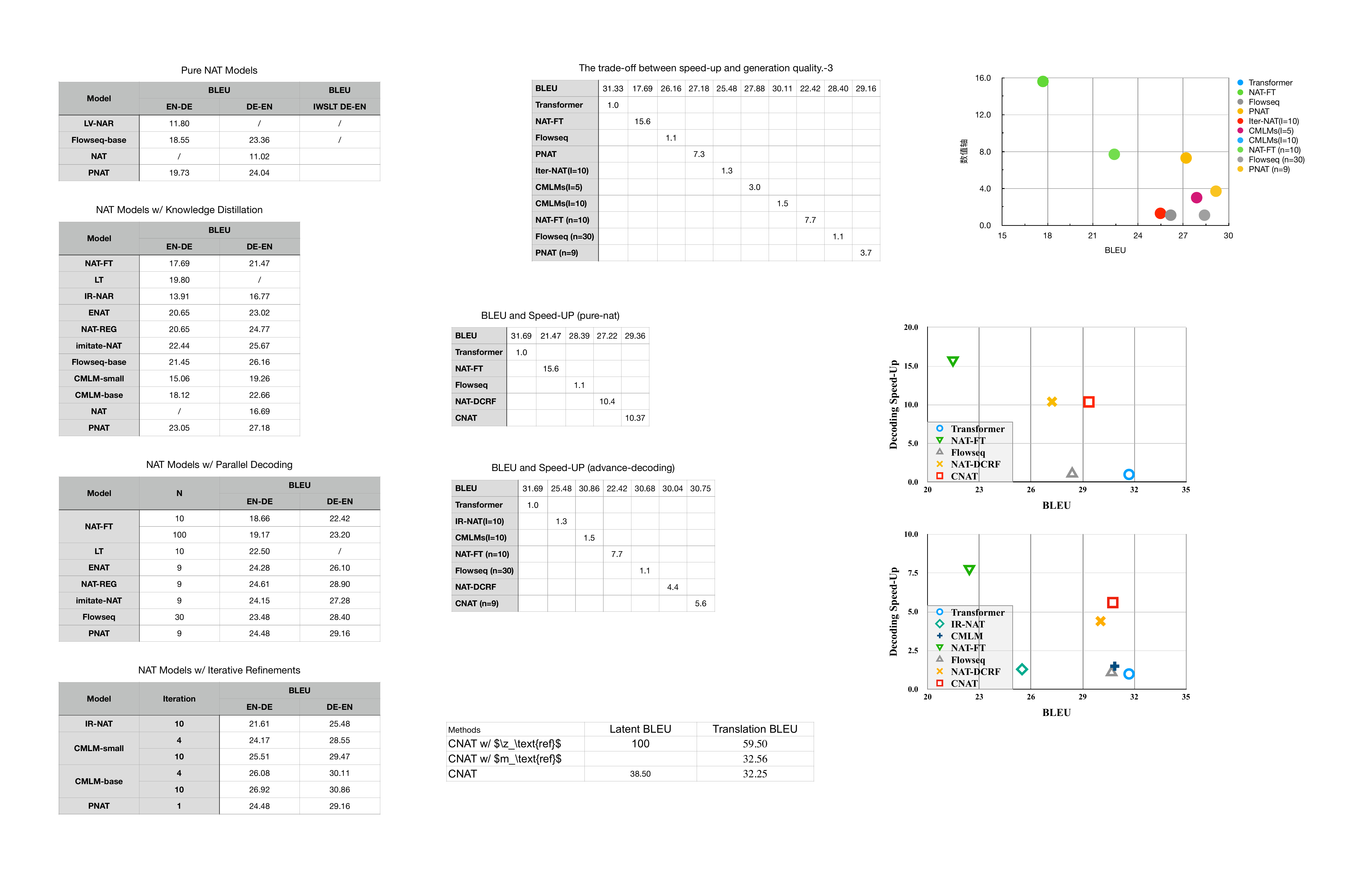}
    \caption{Pure NAT decoding.}
    \label{fig:pure_speed}
    \end{subfigure}
    \begin{subfigure}[t]{0.495\linewidth}
    \includegraphics[width=1.0\linewidth]{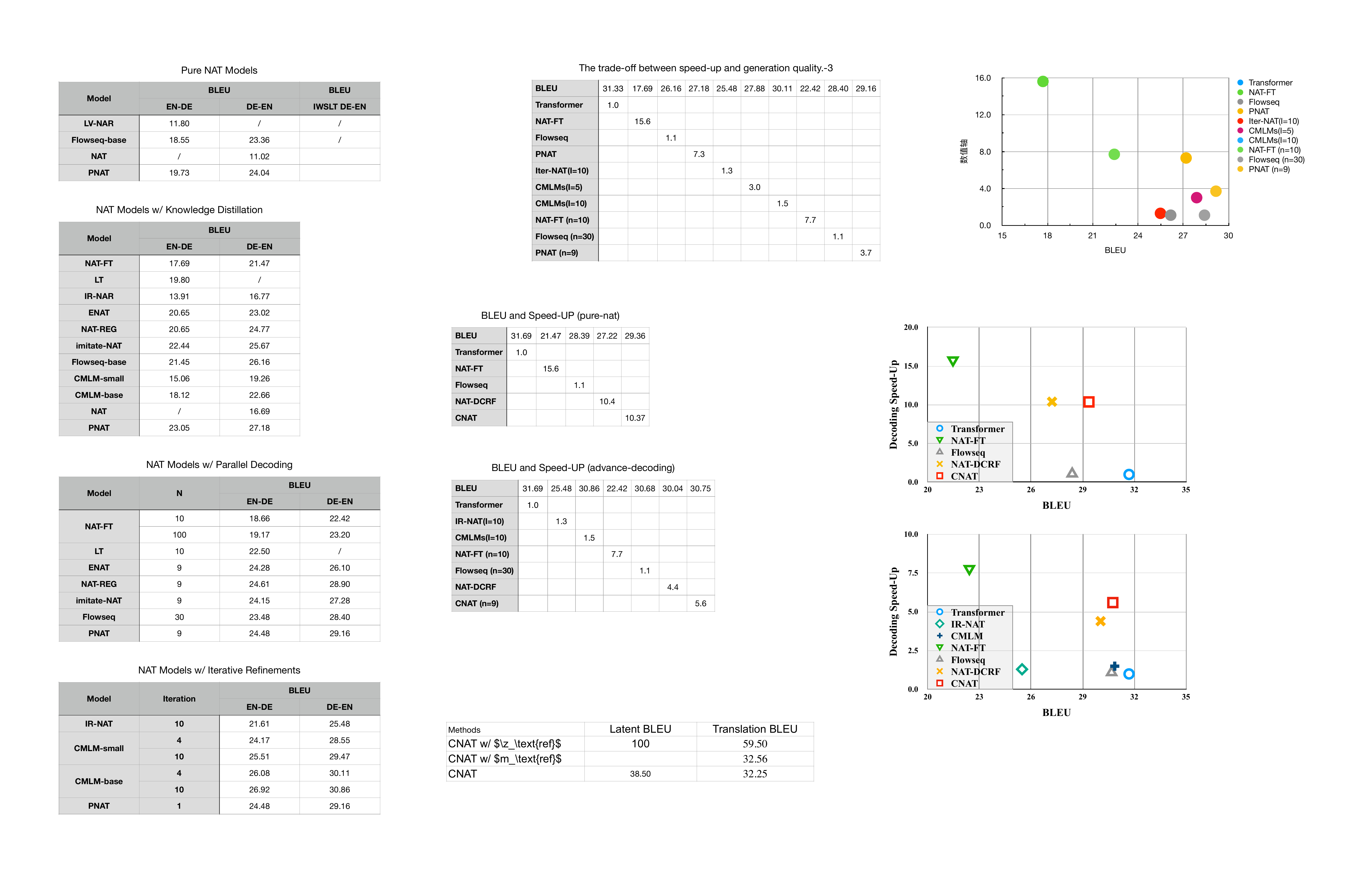}
    \caption{NAT with advanced decoding techniques.}
    \centering
    \label{fig:adv_speed}
\end{subfigure}
\caption{BLEU and decoding speed-up of NAT models on WMT14 DE-EN test set. Each point represents the decoding method run with its corresponding setting in Table~\ref{tab:kd_mt}, Table~\ref{tab:lpd_mt} or Table~\ref{tab:iter_mt}. }
\label{fig:speed}
\end{figure*}
\paragraph{Translation Efficiency.} 
As depicted in Figure~\ref{fig:speed}, we validate the efficiency of~\method. 
Put simply, the decoding speed is measured sentence-by-sentence, and the speed-up is computed by comparing it with the Transformer.
Figure~\ref{fig:pure_speed} and Figure~\ref{fig:adv_speed} show the BLEU scores and decoding speed-up of NAT models. 
The former compares the pure NAT models. 
The latter compares NAT model inference with advanced decoding techniques~(parallel reranking or iterative-based decoding)\footnote{Our results are conducted on a single GeForce GTX 1080-TI GPU. Please note that the result in Figure~\ref{fig:pure_speed} and Figure~\ref{fig:adv_speed} may be evaluated under different hardware settings, and it may not be fair to compare them directly.}.  

We can see from Figure~\ref{fig:speed} that the point of \method is located on the top-right of the baselines. 
The \method outperforms our baselines in BLEU if speed-up is held, and in speed-up if BLEU is held, indicating~\method outperforms previous state-of-the-art NAT methods. 
Although iterative models like CMLM achieves competitive BLEU scores, they only maintain minor speed advantages over Transformer. 
In contrast, \method remarkably improves the inference speed while keeping a competitive performance.

\begin{table}[tbp]
\small
\centering
\begin{tabular}{lcc}
\toprule
Methods                 & Latent BLEU   & Translation BLEU \\
\midrule
\method w/ $\z_\text{ref}$ & 100.00        & 59.12            \\
\method w/ $m_\text{ref}$  & 39.72         & 31.59            \\
\method                    & 38.59         & 31.15           \\ 
\bottomrule
\end{tabular}
\caption{Results on the test of IWSLT14 to analyze the effectiveness of categorical modeling. ``w/ $\z_\text{ref}$'' denote \method generate the tokens condition on the latent sequence which is quantized from the reference target. ``w/ $m_\text{ref}$'' denote the~\method generate the tokens condition on the reference length. }
\label{tab:latent_prediction}
\end{table}
\paragraph{Effectiveness of Categorical Modeling.}
We further conduct the experiments on the test set of IWSLT14 to analyze the effectiveness of our categorical modeling and its influence on translation quality. 
We regard the categorical predictor as a sequence-level generation task and list its BLEU score in Table~\ref{tab:latent_prediction}.

As see, a better latent prediction can yield a better translation. 
With the $\z_\text{ref}$ as the latent sequence, the model achieves surprisingly good performance on this task, showing the usefulness of the learned categorical codes. 
We also can see that the~\method decoding with reference length only slightly ~(0.44 BLEU) better than it with predicted length, indicating that the model is robust.

\begin{table}[tbp]
\centering
\tabcolsep 4pt
\small
\begin{tabular}{ccccccccc}
\toprule
\multirow{2}{*}{Line}&\multicolumn{3}{c}{$K$} && \multicolumn{2}{c}{Predictor} &\multirow{2}{*}{GateNet} & \multirow{2}{*}{BLEU}   \\ \cmidrule{2-4} \cmidrule{6-7}
 &32    & 64    & 128   && CRF         & AR              &                          &  \\\midrule
1&\hit  &       &       && \hit        &                 & \hit                     & 30.13    \\
2&      & \hit  &       && \hit        &                 & \hit                     &\textbf{31.87} \\
3&      &       & \hit  && \hit        &                 & \hit                     & 30.82    \\
4&      & \hit  &       && \hit        &                 &                          & 29.32    \\
5&      & \hit  &       &&             & \hit            & \hit                     & 28.23    \\
6&      & \hit  &       &&             & \hit            &                          & 24.00    \\
7&      &       & \hit  &&             & \hit            &                          & 25.43    \\
8&      &       &       &&             &                 &                          & 24.25    \\
\bottomrule
\end{tabular}
\caption{Ablation study on the dev set of IWSLT14. Note that we train all of the configurations with knowledge distillation. ``AR'' denotes an autoregressive Transformer predictor. The line 8 is our NAT baseline.}
\label{tab:ablation}
\end{table}
\subsection{Ablation Study}
We further conduct the ablation study with different \method variant on dev set of IWSLT14. 

\paragraph{Influence of $K$.}
We can see the CRF with the categorical number $K=64$ achieves the highest score~(line 2).  
A smaller or larger $K$ neither has a better result. 
The AR predictor may have a different tendency: with a larger $K=128$, it achieves a better performance. 
However, a larger $K$ may lead to a higher latency while inference, which is not the best for non-autoregressive decoding.
In our experiments, the $K=64$ can achieve the high-performance and be smaller enough to keep the low-latency during inference. 

\paragraph{CRF versus AR.} 
Experiment results show that the CRF-based predictor is better than the AR predictor. 
We can see that the CRF-based predictor surpasses the Transformer predictor 3.5 BLEU (line 2 vs. line 5) with the GateNet;
without the GateNet, the gap enlarges to 5.3 BLEU (line 4 vs. line 6). 
It is consistent with our intuition that CRF is better than Transformer to model the dependencies among latent variables on machine translation when the number of categories is small. 

\paragraph{GateNet.} 
Without the GateNet, the \method with AR predictor degenerates a standard LT model with a smaller latent space. 
We can see its performance is even lower than the NAT-baselines~(line 6 vs. line 8). 
Equipping with the GateNet and the schedule sampling, it outperforms the NAT baseline with a large margin~(around 4.0 BLEU), showing that the GateNet mechanism plays an essential role in our proposed model.

\subsection{Code Study}
To analyze the learned category, we further compute its relation to two off-the-shelf categorical information: the part-of-speech~(POS) tags and the frequency-based clustered classes.
For the former, we intuitively assign the POS tag of a word to its sub-words and compute the POS tag frequency for the latent codes. 
For the latter, we roughly assign the category of a subword according to its frequency.
It needs to mention that the number of frequency-based classes is the same as that of the POS tags.

\begin{table}[tbp]
\small
\tabcolsep 3pt
\centering
\begin{tabular}{lccc} 
\toprule
               & \textbf{H-score}   & \textbf{C-score}  & \textbf{V-measure}    \\
\midrule
w/ POS tags    & \textbf{0.70}      & 0.47              & \textbf{0.56}         \\
w/ Frequency   & 0.62               &\textbf{0.48}      & 0.54                  \\
\bottomrule
\end{tabular}
\caption{Clustering evaluation metrics on the test set of IWSLT14 to analyze the learned codes. }
\label{tab:cluster}
\end{table}
\paragraph{Quantitative Results.} 
We first compute the V-Measure~\cite{vmeasure} score between the latent categories to POS tags and sub-words frequencies. 
The results are listed in Table~\ref{tab:cluster}.

Overall, the ``w/ POS tags'' achieves a higher V-Measure score, indicating that the latent codes are more related to the POS tags than sub-words frequencies.
The homogeneity score (H-score) evaluates the purity of the category.
We also can see that the former has a relatively higher H-score than the latter (0.70 vs. 0.62),  which is consistent with our intuition.

\begin{figure}[tbp]
\centering
\includegraphics[width=\linewidth]{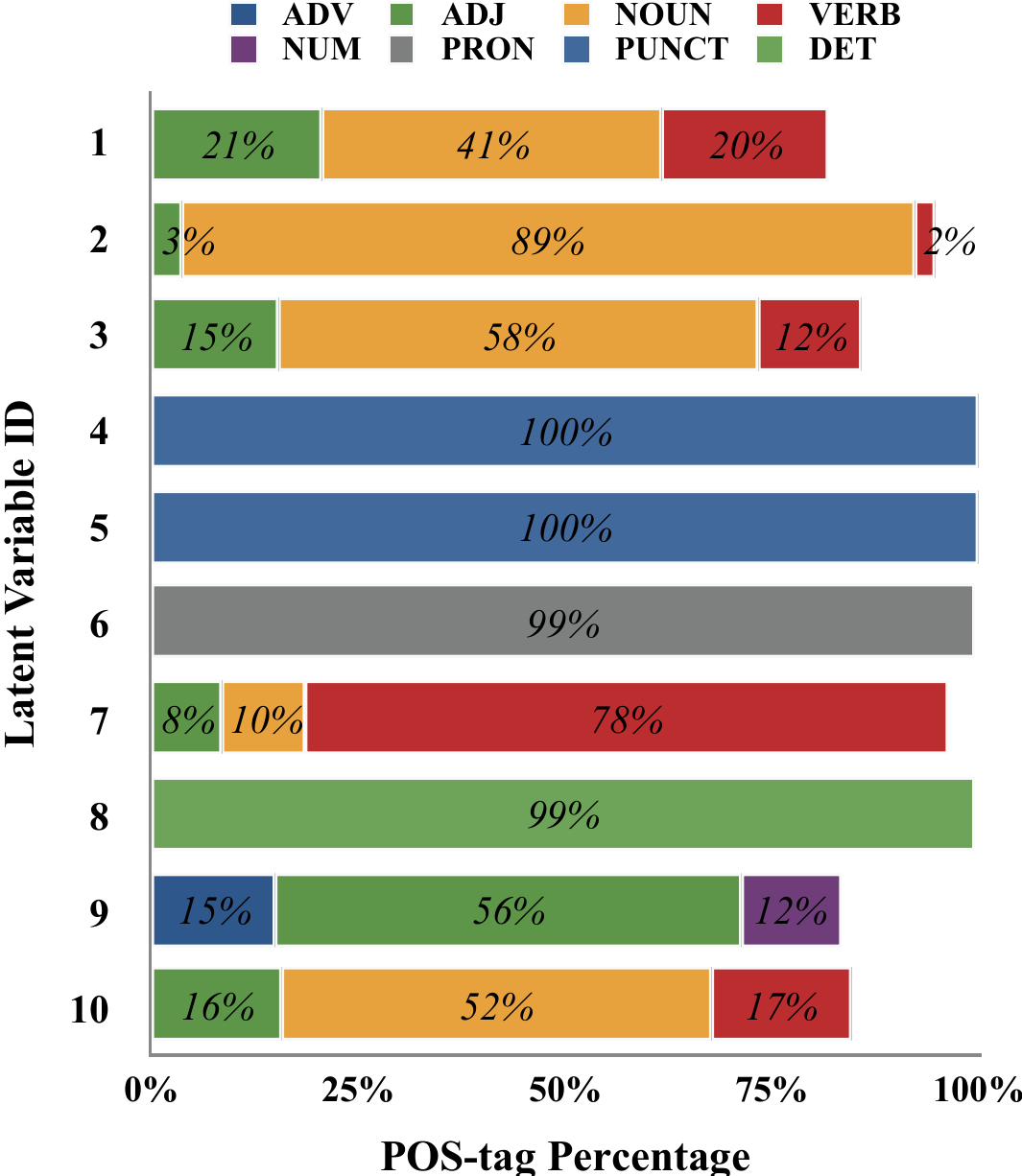}
\caption{The POS tags distribution for the top 10 frequent latent variables on the test set of IWSLT14. We list the top 3 frequent POS tags for each latent variable. }
\label{fig:pos_dist}
\end{figure}
\paragraph{Case Analysis.} 
As shown in Figure~\ref{fig:pos_dist}, we also depict the POS tags distribution for the top 10 frequent latent variables on the test set of IWSLT14\footnote{More details can be found in Appendix B.}. 
We can see a sharp distribution for each latent variable, showing that our learned fuzzy classes are meaningful.

%% file: 05related.tex
\section{Related Work}
\paragraph{Non-autoregressive Machine Translation.}
\citet{nat} first develop a non-autoregressive Transformer~(NAT) for machine translation, which produces the outputs in parallel, and the inference speed is thus significantly boosted. 
Due to the missing of dependencies among the target outputs, the translation quality is largely sacrificed.

A line of work proposes to mitigate such performance degradation by enhancing the decoder inputs.  
\citet{iter_nat} propose a method of iterative refinement based on the previous outputs. 
\citet{enat} enhance decoder input by introducing the phrase table in statistical machine translation and embedding transformation. 
There are also some work focuses on improving the decoder inputs' supervision, including imitation learning from autoregressive models~\citep{imitate_nat} or regularizing the hidden state with backward reconstruction error~\citep{nat_reg}. 

Another work proposes modeling the dependencies among target outputs, which is explicitly missed in the vanilla NAT models. 
\citet{glat,cmlm} propose to model the target-side dependencies with a masked language model, modeling the directed dependencies between the observed target and the unobserved words. 
Different from their work, we model the target-side dependencies in the latent space, which follows the latent variable Transformer fashion.

\paragraph{Latent Variable Transformer.} 
More close to our work is the latent variable Transformer, which takes the latent variable as inputs to modeling the target-side information. 
\citet{lv_nar} combine continuous latent variables and deterministic inference procedure to find the target sequence that maximizes the lower bound to the log-probability. 
\citet{flowseq} propose to use generative flows to the model complex prior distribution. 
\citet{lt} propose to autoregressively decode a shorter latent sequence encoded from the target sentence, then simultaneously generate the sentence from the latent sequence. 
\citet{pnat} model the target position of decode input as a latent variable and introduce a heuristic search algorithm to guide the position learning. 
\citet{syn_st} first autoregressively predict a chunked parse tree and then simultaneously generate the target tokens from the predicted syntax.




%% file: 06conclusion.tex
\section{Conclusion}
We propose~\method, which implicitly models the categorical codes of the target language, narrowing the performance gap between the non-autoregressive decoding and autoregressive decoding.
Specifically, \method builds upon the latent Transformer and models the target-side categorical information with vector quantization and conditional random fields~(CRF) model. 
We further employ a gated neural network to form the decoder inputs. Equipped with the scheduled sampling, \method works more robust.
As a result, the \method achieves a significant improvement and moves closer to the performance of the Transformer on machine translation.

%% file: 07appendix.tex
\section{Non-Indo-European Translation}\label{s:appendix_nist}
\paragraph{Dataset.} 
We apply the \method to the non-Indo-European translation tasks on the LDC Chinese-English\footnote{
LDC2002E18, LDC2003E14, LDC004T08, and LDC2005T06}~(denote as LDC ZH-EN, 1.30M sentence pairs) and MT02 test set of NIST ZH-EN dataset. 
We use \texttt{NLPIRICTCLAS}\footnote{\url{http://ictclas.nlpir.org/}} and Moses tokenizer for Chinese and English tokenization, respectively. 

\begin{table}[htbp]
\centering
\small
\begin{tabular}{lc}
\toprule
Model           & BLEU \\
\midrule
Transformer     &  28.05    \\
NAT             &  12.31    \\
\method         &  \textbf{22.16}    \\
\bottomrule
\end{tabular}
\caption{Results on the MT02 set of different models.}
\label{tab:nist}
\end{table}
\paragraph{Results.} 
We can see than in Table~\ref{tab:nist} that our model can enhance the performance of NAT with a large margin~(22.16 vs. 12.31).

\section{Learned Latent Codes}\label{sec:appendix_code}

\begin{table}[htbp]
\centering
\small
\tabcolsep 2pt
\begin{tabular}{lrrr}
\toprule
ID & Top 1 & Top 2 & Top 3\\
\midrule
0& NOUN(41.06\%)& ADJ(20.74\%)& VERB(20.09\%)\\
1& NOUN(88.73\%)& ADJ(3.48\%)& VERB(2.34\%)\\
2& NOUN(58.06\%)& ADJ(15.40\%)& VERB(12.29\%)\\
3& PUNCT(100.00\%)& ---& ---\\
4& PUNCT(99.80\%)& NOUN(0.14\%)& PROPN(0.05\%)\\
5& PRON(99.29\%)& NOUN(0.42\%)& PROPN(0.20\%)\\
6& VERB(77.95\%)& NOUN(9.93\%)& ADJ(8.45\%)\\
7& DET(99.33\%)& PUNCT(0.47\%)& PART(0.15\%)\\
8& ADJ(56.24\%)& ADV(15.06\%)& NUM(11.87\%)\\
9& NOUN(52.01\%)& VERB(16.72\%)& ADJ(15.75\%)\\
10& CCONJ(99.71\%)& VERB(0.11\%)& NOUN(0.11\%)\\
11& PART(71.00\%)& ADP(27.70\%)& SCONJ(0.99\%)\\
12& PRON(60.71\%)& SCONJ(30.15\%)& DET(8.88\%)\\
13& ADP(90.70\%)& SCONJ(5.91\%)& ADV(3.09\%)\\
14& DET(96.81\%)& NOUN(1.81\%)& PROPN(1.13\%)\\
15& NOUN(88.49\%)& VERB(4.28\%)& ADV(3.10\%)\\
16& VERB(90.35\%)& NOUN(4.82\%)& AUX(4.40\%)\\
17& ADV(67.37\%)& PART(15.02\%)& NOUN(8.31\%)\\
18& PRON(68.46\%)& DET(29.31\%)& NOUN(1.65\%)\\
19& ADV(61.05\%)& SCONJ(36.01\%)& ADP(2.31\%)\\
20& VERB(93.70\%)& NOUN(6.03\%)& PROPN(0.23\%)\\
21& PRON(98.59\%)& NOUN(0.78\%)& ADJ(0.39\%)\\
22& ADP(93.45\%)& ADV(2.40\%)& SCONJ(1.95\%)\\
23& AUX(85.14\%)& VERB(14.51\%)& NOUN(0.36\%)\\
24& AUX(80.60\%)& VERB(18.55\%)& NOUN(0.60\%)\\
25& VERB(44.78\%)& AUX(40.84\%)& PROPN(13.49\%)\\
26& ADP(73.68\%)& ADV(9.68\%)& SCONJ(9.55\%)\\
27& AUX(99.47\%)& NOUN(0.53\%)& ---\\
28& AUX(76.11\%)& VERB(10.26\%)& PART(9.70\%)\\
29& ADP(89.48\%)& ADV(3.93\%)& NOUN(3.93\%)\\
30& VERB(43.34\%)& ADP(34.41\%)& AUX(14.47\%)\\
31& DET(53.56\%)& PRON(46.44\%)& ---\\
32& ADV(95.89\%)& SCONJ(3.72\%)& PROPN(0.23\%)\\
33& ADP(52.12\%)& SCONJ(27.84\%)& ADV(8.74\%)\\
34& DET(62.63\%)& PRON(20.73\%)& ADV(9.90\%)\\
35& VERB(76.34\%)& AUX(23.66\%)& ---\\
36& AUX(100.00\%)& ---& ---\\
37& ADV(47.90\%)& PRON(40.51\%)& NOUN(11.59\%)\\
38& NOUN(99.17\%)& ADJ(0.73\%)& ADV(0.10\%)\\
39& NOUN(35.74\%)& VERB(28.94\%)& ADJ(28.21\%)\\
40& NOUN(49.84\%)& ADJ(27.23\%)& ADV(16.19\%)\\
41& PRON(92.22\%)& NOUN(4.75\%)& PROPN(1.08\%)\\
42& PRON(92.73\%)& DET(7.05\%)& ADV(0.22\%)\\
43& ADP(32.17\%)& NOUN(23.05\%)& ADJ(21.61\%)\\
44& PUNCT(100.00\%)& ---& ---\\
45& PRON(100.00\%)& ---& ---\\
46& AUX(94.36\%)& VERB(4.59\%)& NOUN(1.05\%)\\
47& ADV(67.33\%)& ADJ(31.75\%)& NOUN(0.79\%)\\
48& ADP(91.70\%)& SCONJ(8.16\%)& ADJ(0.14\%)\\
49& PUNCT(100.00\%)& ---& ---\\
50& PART(70.91\%)& AUX(25.04\%)& ADP(2.85\%)\\
51& CCONJ(99.52\%)& ADP(0.48\%)& ---\\
52& ADP(69.34\%)& SCONJ(15.89\%)& ADV(14.13\%)\\
53& PUNCT(100.00\%)& ---& ---\\
54& NOUN(58.00\%)& VERB(30.58\%)& ADJ(6.15\%)\\
55& VERB(71.57\%)& AUX(28.04\%)& NOUN(0.39\%)\\
56& NUM(75.73\%)& NOUN(20.33\%)& PRON(2.49\%)\\
57& DET(86.03\%)& ADJ(6.77\%)& PROPN(3.71\%)\\
58& ADP(61.07\%)& ADV(31.77\%)& NOUN(5.37\%)\\
59& CCONJ(90.75\%)& NOUN(8.48\%)& ADJ(0.51\%)\\
60& ADP(78.74\%)& SCONJ(19.93\%)& ADV(1.00\%)\\
61& VERB(59.11\%)& NOUN(31.56\%)& ADJ(9.33\%)\\ 
\bottomrule
\end{tabular}
\caption{The distribution of pos tags for latent variables. For each latent variable, we list the top 3 frequent pos tags and their corresponding percentages. }
\end{table}